\documentclass[11pt,twoside]{article}

\usepackage{asp2014}

\aspSuppressVolSlug
\resetcounters

\begin{document}

\title{Morphological classification of astronomical images with limited labelling}

\author{Andrey~Soroka,$^1$ Alex~Meshcheryakov,$^2$ and Sergey~Gerasimov$^1$}
\affil{$^1$Moscow State University CMC, Moscow, Russia; \email{soroka_irk@mail.ru}}
\affil{$^2$Space Research Institute of RAS, Moscow, Russia;
\email{mesch@cosmos.ru}}

\paperauthor{Andrey~Soroka}{soroka_irk@mail.ru}{}{Moscow State University}{Computational Mathematics and Cybernetics}{Moscow}{}{119234}{Russia}
\paperauthor{Alex~Meshcheryakov}{mesch@cosmos.ru}{}{Russian academy of sciences}{Space Research Institute}{Moscow}{}{117997}{Russia}
\paperauthor{Sergey~Gerasimov}{sergun@gmail.com}{}{Moscow State University}{Computational Mathematics and Cybernetics}{Moscow}{}{119234}{Russia}



\begin{abstract}
The task of morphological classification is complex for simple parameterization, but important for research in the galaxy evolution field. Future galaxy surveys (e.g. EUCLID) will collect data about more than a $10^9$ galaxies. To obtain morphological information one needs to involve people to mark up galaxy images, which requires either a considerable amount of money or a huge number of volunteers. We propose an effective semi-supervised approach for galaxy morphology classification task, based on active learning of adversarial autoencoder (AAE) model. For a binary classification problem (top level question of Galaxy Zoo 2 decision tree) we achieved accuracy 93.1\% on the test part with only 0.86 millions markup actions, this model can easily scale up on any number of images. Our best model with additional markup achieves accuracy of 95.5\%. To  the  best  of  our  knowledge  it is a  first time AAE semi-supervised learning model used in astronomy.
\end{abstract}



\section{Introduction}
Galaxies exhibit a wide variety of shapes, colors and sizes. These properties are indicative of their age, formation conditions, and interactions with other galaxies over the course of many Gyr. Studies of galaxy formation and evolution use morphology to probe the physical processes that give rise to them. Traditionally, morphological classification has been carried out by professional astronomers using visual inspections of images obtained from telescopes. The number of detected galaxies in modern sky surveys and computer simulations has been found to be growing rapidly. The number of images in surveys (SDSS, DESI LIS, PanSTARRS) is billions of galaxies, future surveys such as Euclid \citep{refregier2010euclid} will provide images of more than billion of galaxies. Manually inspecting all these images to annotate them with morphological information is impractical for either individual astronomers or small teams.

An attempt to solve this problem was the Galaxy Zoo project by \citet{Willett_2013}: volunteers classified images of galaxies through a web interface. Participants were asked various questions such as "How rounded is the galaxy?" or "Is there a convex in the center?" Depending on the answer, it was determined which question would be asked next. The questions formed a decision tree that covered a wide range of morphological categories. The crowdsourcing project also faces the problem of a large amount of data: if we try to study the morphology of only 1\% of the volume of images from Euclid, we will get millions of objects for morphological classification. A project of this magnitude would take over sixty years to process all the data at the pace of the Galaxy Zoo 2.

There are approaches to galaxy classification based on the ideas of active learning. They allow to increase the classification speed by ~8 times at the expense of a slight loss of quality, but even at this marking rate, it will take \textbf{7.5} years for the SWAP by \citet{Beck_2018} (SoTA model) to handle 1\% of all Euclid data.

The main objectives of our research: 1) create a model for the morphological classification of galaxies, 2) the model should use small as possible amount of labeled data, 3) predictive ability should be better than current SoTA solutions, 4) use of active learning technique. The second task is the main one, since reducing the number of required markup images will reduce the number of primary \textbf{classification actions} carried out by volunteers / research groups. This, in turn, will significantly speed up the marking of morphological annotations of modern surveys.

\section{Data}
We chose Kaggle Galaxy Zoo dataset for our experiments. With 138000 images of galaxies in jpg format. As mentioned above, Galaxy Zoo is a crowdsourcing project, each image in it was classified based on the votes of volunteers, an average of 42 people per galaxy. Thus, we have a dependence between amount of labeled data and markup actions: 1 labeled image $\approx42$  markup actions. Annotation of data was provided by Kyle Willett, KGZ organizer. Authors are grateful for his help and kind words.

\section{Methods}
In this work, we propose semi-supervised algorithm, modernization of the adversarial autoencoder by \citet{makhzani2016adversarial}. We did classification over a continuous latent representation using additional MLP layer. We trained encoder in semi-supervised mode with both labeled and unlabeled data, the second was an order of magnitude more. In the process of training adversarial autoencoder, in addition to the classical "uniform" selection of training samples, we also used active learning approach. The training examples were selected sequentially, based on the prediction of the model trained in the previous step with fewer examples. At the first step, 4\% of the total number of images were randomly marked as a training sample, model was trained on them, other images were classified with that model, and the next 1\% of the total number of images was marked, for which algorithm predicted the probability closest to 0.5 (random prediction). The model was retrained, process continued until the total amount of labeled data was equal to 10\% of all training data. We approximated the results to the dataset size used by \citet{Beck_2018}. To do this, we took the accuracy obtained on the unlabeled part and transferred it to an enlarged one in accordance with the size of the SWAP dataset minus 20k labeled images and calculated the final results as a composition of these values.

\articlefigure[width=.6\textwidth]{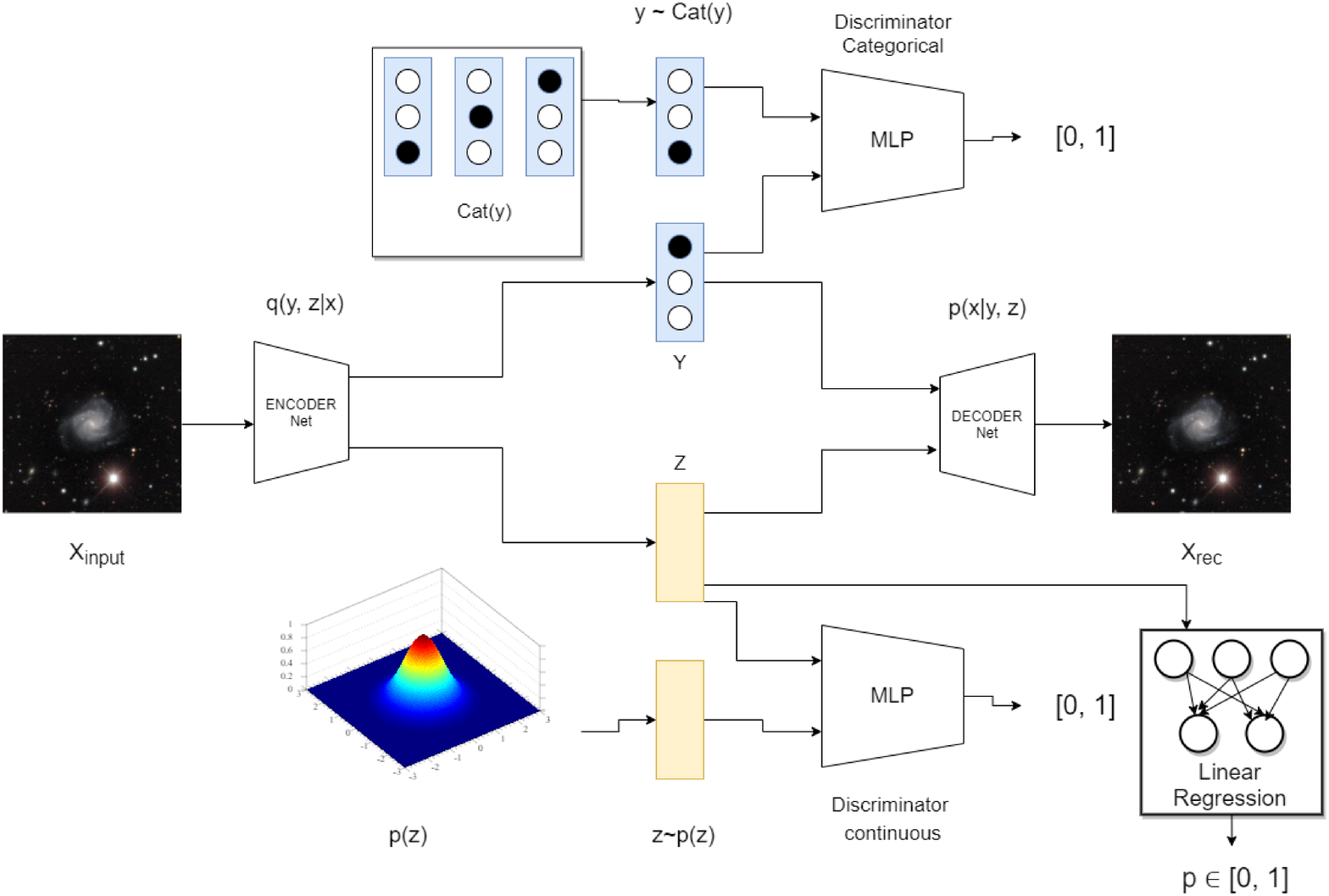}{fig1}{Architecture of semi-supervised AAE model.}

\section{Results}
The main learning outcomes: 1) Semi-supervised model (AAE) achieves an accuracy of 91\% on 10\% labeled data ($\sim$850,000 markup actions for 20k images) of all training data, which is quite close to the result of a similar SWAP model (93.1\% on $\sim$936000 markup actions); 2) Active learning improves prediction, AAE+AL model trained on the same amount of labeled images achieves an accuracy of 93.1\%. This result can be easily scaled up on any amount of images without additional markup cost; 3) If we estimate the accuracy over the entire Galaxy Zoo 2 dataset (226124 images as in \citet{Beck_2018}), including labeled examples (as the \citet{Beck_2018} of SWAP did ), our model achieves an accuracy of 95.5\%, which is quite close to the best SWAP result (95.7\%). At the same time, it requires 2.5 times less markup actions ($\sim$850000 vs. $\sim$2300000 for SWAP).

\articlefigure[width=.55\textwidth]{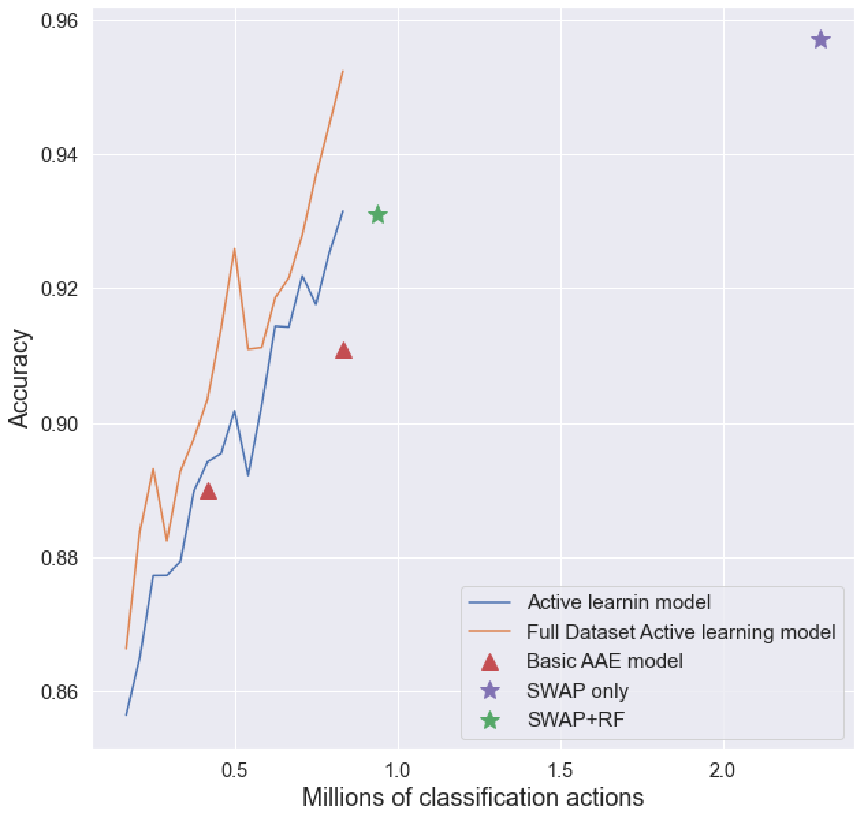}{fig2}{Accuracy of different models over active learning process.}

\begin{table}[!ht]
\caption{Accuracy of different models over active learning process. All AAE models are from this work and all SWAP models are from \citet{Beck_2018}}
\smallskip
\begin{center}
{\small
\begin{tabular}{lllll}  
\tableline
\noalign{\smallskip}
Model & Acc.(\%) & Galaxies & Markup actions & Add markup? \\
\noalign{\smallskip}
\tableline
\noalign{\smallskip}
SWAP  & \textbf{95.7} & 226124 & 2298772 & Yes \\
AAE+AL  & \textbf{95.5} & 226124 & \textbf{854310} & Yes \\
SWAP+RF  & 93.1 & 210803 & 936887 & Yes \\
AAE+AL (20k sample) & 93.1 & \textbf{Any} & 854310 & \textbf{No}
\end{tabular}
}
\end{center}
\end{table}

Let $N$ be the size of the entire dataset to be classified, A be the limit of labeled images used in active learning. Let's denote by $R$ - the ratio of $N$ to $A$: $R = \frac{N}{A}$. We measured how the prediction accuracy of the model changes with a fixed $A$ and growing $N$. We sampled data with the $N$ range from 30k to 130k with $A$ equal to 7k and 20k. We note that the best accuracy of classification is achieved with more labeled data. So if we increase the total amount of data, but keep the value R, we get the best results (Figures \ref{fig3}).  

\articlefigure[width=.9\textwidth]{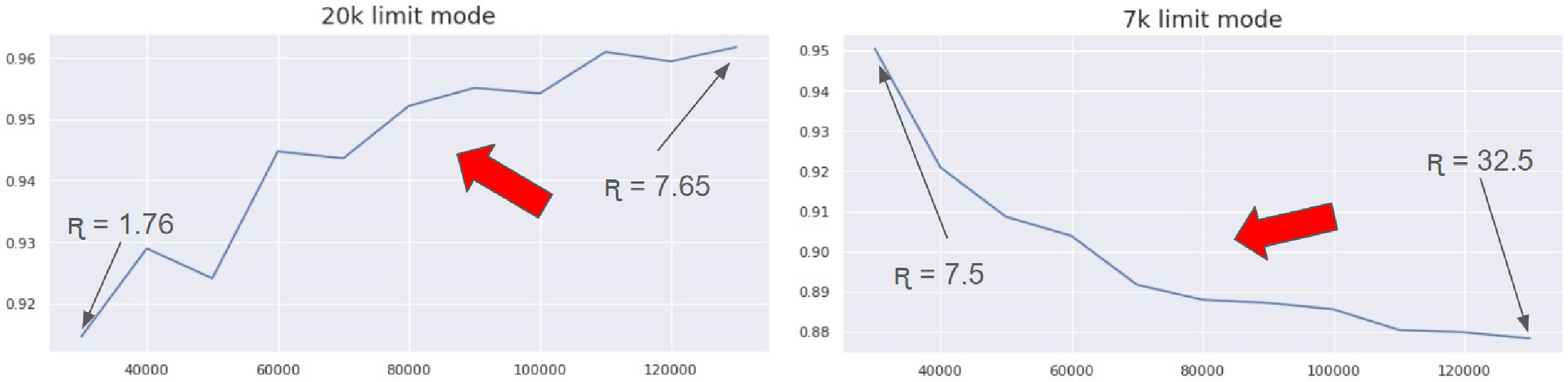}{fig3}{Accuracy of different models over active learning process.}

\section{Conclusion}
To the best of our knowledge its a first time AAE used in astronomy. Our model achieves accuracy 93.1\% on 20k labeled data without additional markup cost, this model can easily scale up on \textbf{any number of images}. Our best model with additional markup achieves accuracy of 95.5\%, while using \textbf{2.5 times} less markup actions than \citet{Beck_2018}.


\bibliography{P7-206}  

\begin{thebibliography}{}
\expandafter\ifx\csname natexlab\endcsname\relax\def\natexlab#1{#1}\fi
\expandafter\ifx\csname url\endcsname\relax
  \def\url#1{\texttt{#1}}\fi
\expandafter\ifx\csname urlprefix\endcsname\relax\def\urlprefix{URL }\fi
\providecommand{\eprint}[2][]{\url{#2}}

\bibitem[{Beck et~al.(2018)Beck, Scarlata, Fortson, Lintott, Simmons, Galloway,
  Willett, Dickinson, Masters, Marshall, \& et~al.}]{Beck_2018}
Beck, M.~R., Scarlata, C., Fortson, L.~F., Lintott, C.~J., Simmons, B.~D.,
  Galloway, M.~A., Willett, K.~W., Dickinson, H., Masters, K.~L., Marshall,
  P.~J., \& et~al. 2018, Monthly Notices of the Royal Astronomical Society,
  476, 5516–5534. \urlprefix\url{http://dx.doi.org/10.1093/mnras/sty503}

\bibitem[{Makhzani et~al.(2016)Makhzani, Shlens, Jaitly, Goodfellow, \&
  Frey}]{makhzani2016adversarial}
Makhzani, A., Shlens, J., Jaitly, N., Goodfellow, I., \& Frey, B. 2016,
  Adversarial autoencoders. \eprint{1511.05644}

\bibitem[{Refregier et~al.(2010)Refregier, Amara, Kitching, Rassat, Scaramella,
  \& Weller}]{refregier2010euclid}
Refregier, A., Amara, A., Kitching, T.~D., Rassat, A., Scaramella, R., \&
  Weller, J. 2010, Euclid imaging consortium science book. \eprint{1001.0061}

\bibitem[{Willett et~al.(2013)Willett, Lintott, Bamford, Masters, Simmons,
  Casteels, Edmondson, Fortson, Kaviraj, Keel, \& et~al.}]{Willett_2013}
Willett, K.~W., Lintott, C.~J., Bamford, S.~P., Masters, K.~L., Simmons, B.~D.,
  Casteels, K. R.~V., Edmondson, E.~M., Fortson, L.~F., Kaviraj, S., Keel,
  W.~C., \& et~al. 2013, Monthly Notices of the Royal Astronomical Society,
  435, 2835–2860. \urlprefix\url{http://dx.doi.org/10.1093/mnras/stt1458}

\end{thebibliography}


\end{document}